\documentclass[sigconf, nonacm]{acmart}

\AtBeginDocument{%
  }

\acmISBN{978-1-4503-XXXX-X/18/06}

\usepackage[utf8]{inputenc}
\usepackage{algorithm}
\usepackage{algpseudocode}
\usepackage{multirow}
\usepackage{graphicx}
\usepackage{subcaption}
\usepackage{pifont}
\usepackage{caption}
\usepackage{subcaption}

\algrenewcommand\alglinenumber[1]{\normalsize #1:}
\algrenewcommand\algorithmicrequire{\textbf{Input:}}
\algrenewcommand\algorithmicensure{\textbf{Output:}}
\algnewcommand{\LeftComment}[1]{\(\triangleright\) #1}



\begin{document}

\title{ARES: Auxiliary Range Expansion for Outlier Synthesis}

\author{Eui-Soo Jung}
\affiliation{
  \institution{University of Seoul}
  \city{Seoul}
  \country{Republic of Korea}
}
\email{jyssys12@uos.ac.kr}

\author{Hae-Hun Seo}
\affiliation{
  \institution{University of Seoul}
  \city{Seoul}
  \country{Republic of Korea}
}
\email{hun@uos.ac.kr}

\author{Hyun-Woo Jung}
\affiliation{
  \institution{University of Seoul}
  \city{Seoul}
  \country{Republic of Korea}
}
\email{nero0115@uos.ac.kr}

\author{Je-Geon Oh}
\affiliation{
  \institution{University of Seoul}
  \city{Seoul}
  \country{Republic of Korea}
}
\email{kletaln4792@uos.ac.kr}

\author{Yoon-Yeong Kim}
\authornotemark[0]
\affiliation{%
  \institution{University of Seoul}
  \city{Seoul}
  \country{Republic of Korea}
}
\email{yykim@uos.ac.kr}
\authornote{Corresponding author.}








\renewcommand{\shortauthors}{Eui-Soo Jung et al.}

\begin{abstract}
Recent successes of artificial intelligence and deep learning often depend on the well-collected training dataset which is assumed to have an identical distribution with the test dataset. However, this assumption, which is called closed-set learning, is hard to meet in realistic scenarios for deploying deep learning models. As one of the solutions to mitigate this assumption, research on out-of-distribution (OOD) detection has been actively explored in various domains. In OOD detection, we assume that we are given the data of a new class that was not seen in the training phase, i.e., outlier, at the evaluation phase. The ultimate goal of OOD detection is to detect and classify such unseen outlier data as a novel "unknown" class. Among various research branches for OOD detection, generating a virtual outlier during the training phase has been proposed. However, conventional generation-based methodologies utilize in-distribution training dataset to imitate outlier instances, which limits the quality of the synthesized virtual outlier instance itself. In this paper, we propose a novel methodology for OOD detection named Auxiliary Range Expansion for Outlier Synthesis, or ARES. ARES models the region for generating out-of-distribution instances by escaping from the given in-distribution region; instead of remaining near the boundary of in-distribution region. Various stages consists ARES to ultimately generate valuable OOD-like virtual instances. The energy score-based discriminator is then trained to effectively separate in-distribution data and outlier data. Quantitative experiments on broad settings show the improvement of performance by our method, and qualitative results provide logical explanations of the mechanism behind it.
\end{abstract}



\keywords{Out-of-Distribution Detection, Mixup, Auxiliary Distribution, Energy-based Score}


\maketitle

\section{Introduction} \label{sec:Introduction}

The big data era has opened the blooming of artificial intelligence and deep learning. A large set of collected data and development of learning algorithms are being applied in various domains, including computer vision \cite{ResNet, YOLO, U_Net, Diffussion}, natural language processing \cite{BERT, GPT3}, signal processing \cite{Jukebox, WaveNet}, etc.
However, most of the successes of deep learning have been accomplished in a closed-set learning scenario where we assume that both the training dataset and the test dataset follow the same distribution. This assumption leads the deep learning models to solely focus on effectively extracting the hidden representation from the given training dataset.

In the real-world setting, however, this assumption is hard to meet in various deployment phases. That is, the distribution of the training dataset and the test dataset may differ, which is called as open-set learning scenario.
Various research branches have been developed to treat this realistic setting by slightly varying the application scenarios; including novelty detection \cite{SVM}, anomaly detection \cite{Deep_One-Class}, open-set recognition \cite{OpenMax}, etc.

Out-of-distribution (OOD) detection \cite{MSP} is one such variant for open-set learning. In the OOD detection problem, we are given a training dataset as the resource for learning the model; and instances of unknown classes, which were unseen in the training dataset, are included in the test dataset.
In this problem, we denote the training instances as the in-distribution dataset, while we denote the unseen instances as the out-of-distribution dataset.
Having said that, the ultimate goal of the OOD detection is to recognize the instances of the unseen class and label the instances as a novel "unknown" class (i.e., out-of-distribution instance), instead of mislabeling the instance as one of the known classes (i.e., in-distribution instance) that belong to the training dataset. Simultaneously, the model should also predict the ground-truth label of the in-distribution dataset accurately.

Focusing on the developments of the OOD detection methodologies, various branches of methodology have been proposed so far.
One of the most simple but effective branches is the classifier-based method. This branch of methods uses the confidence value of the softmax output produced by the model, and labels each instance as out-of-distribution if the confidence value for the instance does not exceed a threshold.
Moreover, energy score-based methods have also been proposed. In these branches of methods, an additional discriminator is introduced to discriminate in-distribution and out-of-distribution instances using energy-based score \cite{Energy_Score}. The introduction of the discriminator brings some advantages in that it does not harm the discriminative representation learned by the classifier model while effectively screening out the out-of-distribution instances.
Recently, data generation-based methods are also being actively explored, adopting Generative Adversarial Networks (GAN) \cite{GAN_Ian} or data augmentations \cite{ImageNet_CNN}. These branches generate virtual instances that act as probable out-of-distribution instances, or outlier instances. The generated virtual outlier instances are further used for calculating the energy-based score that is distinguished from the given training dataset.
Overall, these methods are jointly used in diverse directions, leading to an improved performance of OOD detection.

However, there exist some limitations of data generation-based methods.
First, these methods sometimes require additional effort to generate the virtual outlier. For example, adopting GAN takes computational cost and elaborate effort for training the generative model until its convergence. Moreover, the quality of the generated virtual outlier instances depends on the training quality of GAN.
Second, the generation process starts with the given training dataset, i.e., in-distribution dataset. Hence, the generated virtual outlier instances are located near the boundary of in-distribution region, and these instances do not sufficiently exhibit the characteristics of true out-of-distribution dataset.

This paper aims to overcome the limitations of the conventional data generation-based methods above.
First, we need to adopt a simple but effective generation process for virtual outliers, rather than training an additional network. Second, we need to get out of the in-distribution region and generate virtual outlier instances far away from the in-distribution region.
These two views lead us to the proposal of our novel algorithm named \underline{A}uxiliary \underline{R}ange \underline{E}xpansion for Outlier \underline{S}ynthesis (ARES). ARES uses a sampling-based method to generate virtual outliers, which does not require additional training cost or effort. The sampling of virtual outliers is conducted in the estimated out-of-distribution space that is expanded far away from the given in-distribution region. In specific, our virtual out-of-distribution space is estimated from the Mixup data \cite{Mixup, Manifold_Mixup}.

The remainder of this paper is as follows.
Section \ref{sec:Background} covers the background knowledge for the development of our method. Section \ref{sec:Method} introduces the motivation of our method, ARES, and treats the detailed algorithms. Section \ref{sec:Experiment} shows the improved performance of OOD detection by using ARES and Section \ref{sec:Conclusion} follows to conclude this paper.

\section{Background} \label{sec:Background}


\subsection{Mixup-based Augmentation}
Data augmentation is a widely used technique for enhancing the performance of deep learning models \cite{ImageNet_CNN}. Traditional data augmentation includes geometric transformations such as rotations, horizontal/vertical flipping, cropping, etc. \cite{Going_deeper, Densely_connected_Conv}.

Recent advancements in data augmentation involve Mixup-based techniques that address the issues of ERM (Empirical Risk Minimization) \cite{Mixup}. Unlike ERM, which trains the model with the original training dataset, Mixup trains the model by mixing two data instances and their labels with a mixing coefficient, $\lambda$.
Here, $\lambda$ is sampled from $Beta(\alpha, \alpha)$ distribution, where $\alpha \in (0, \infty)$; resulting in a mixed instance, $\tilde{x} = \lambda x_{i} + (1-\lambda) x_{j}$, and its mixed label, $\tilde{y} = \lambda y_{i} + (1-\lambda) y_{j}$. Moreover, Manifold Mixup has been proposed to mix feature maps that are randomly selected, instead of raw input images, to learn a smoother decision boundary \cite{Manifold_Mixup}.

Recently, the Mixup-based methods have been explored for enhancing the performance of deep learning models in various areas. 
PixMix is one such development of Mixup, which deals with Machine Learning safety including calibration, anomaly detection, corruption, consistency, and adversaries \cite{PixMix}. PixMix mixes original data with auxiliary data such as fractal images to utilize the natural structural complexity of pictures.


\subsection{Energy-based Score}
An energy-based model maps $d$-dimensional inputs to a non-probabilistic scalar energy, which is called energy-based score.
Energy-based score is widely adopted in OOD detection models as an indicator to discriminate between in-distribution and out-of-distribution instances.
Energy-based score is advantageous in that it resolves the issue of arbitrarily high values of softmax confidence of OOD instances, leading to an effective discrimination of ID and OOD instances \cite{Energy_Score}.

\subsection{Data Generation-based OOD Detection}
In data generation-based out-of-distribution detection methods, virtual out-of-distribution instances are generated from the in-distribution dataset during the training process.
By introducing an additional loss to discriminate the virtual instances from the original instances, the model can learn the representations of the out-of-distribution dataset during the training phase.

Data augmentation has been widely used for the generation of virtual OOD instances.
Contrastive Shifted Instances (CSI) adopt hard augmentation (e.g., rotation) to produce a negative pair, and the augmented instances act as virtual OOD instances \cite{CSI}.
However, using traditional data augmentations also limits the quality of virtual OOD instances in that they are generated near the boundary of the in-distribution region and may not contain a sufficient representation of true OOD instances.

Mixture Outlier Exposure (MixOE) utilizes an auxiliary dataset for generating virtual instances \cite{MixOE}. In MixOE, virtual OOD instances are generated in the form of both fine-grained and coarse-grained by mixing ID instances with real OOD instances.
However, the auxiliary dataset used for MixOE requires a heavy real OOD dataset, and it may appear as if the model is learning from actual OOD samples.

GAN-based methods adopt Generative Adversarial Network (GAN) \cite{GAN_Ian} to generate realistic virtual instances.
In this line of research, the Kullback-Leibler divergence term serves to discriminate virtual instances as OOD instances by approximating a uniform distribution for the virtual instances \cite{GAN-based_OOD}.
However, synthesizing data in high-dimensional feature space is challenging to optimize; and due to the nature of GAN, it requires high computational costs as well as an increased training time.

Recently, sampling-based methods have been proposed as a simple but effective way for generating OOD instances to overcome the burden of adopting GAN.
Virtual Outlier Synthesis (VOS) estimates the distribution of the training dataset as a multivariate Gaussian distribution of $\mathcal{N}(\mu,\sigma)$ \cite{du2021vos}. Then, VOS samples the virtual OOD instances of a specific class $k$, which is denoted as $\mathbf{v}_k$, from the farthest region from the training distribution, i.e., the region where the likelihood in the feature space is the lowest. The sampling process is implemented in a more manageable low-dimensional feature space, rather than the high-dimensional feature space that GAN-based methods use.
These virtual OOD instances are then trained along with in-distribution data to present an uncertainty loss based on an energy-based score \cite{Energy_Score}.

Based on VOS, Reliable Outlier Synthesis under Noisy Feature Space (RONF) has been proposed recently \cite{RONF}. In RONF, out-of-distribution instances generated by VOS are newly mixed by Mixup among them. This mixing process of virtual OOD instances naturally leads them to lie in the out-of-distribution region. RONF also introduces a new scoring function named energy with energy discrepancy score, based on the existing energy-based score.

Non-parametric Outlier Synthesis (NPOS) is also a methodology based on VOS \cite{NPOS}. Unlike VOS, which estimates the feature space distribution parametrically as a multivariate Gaussian distribution, NPOS estimates Gaussian distribution based on the ID embeddings, which are collected at the boundary based on the non-parametric k-NN distances. Then, virtual outliers are synthesized with the lowest likelihood based on rejection sampling \cite{rejection_sampling}. However, methodologies based on such non-parametric estimation tend to incur relatively higher computational and time costs; and are also dependent on well-performing embedders like CLIP \cite{CLIP}.

Additionally, conventional sampling-based methods above synthesize OOD instances that do not deviate significantly from the in-distribution boundary. This means that the synthesized instances do not greatly differ from instances belonging to the in-distribution. Our proposed methodology, named ARES, is built upon the sampling-based method like VOS, but ARES differentiates itself in that it expands the OOD region where virtual instances are sampled.
In a nutshell, ARES extends the estimation of out-of-distribution beyond the in-distribution boundary and synthesizes OOD instances from the broader range.




\begin{figure*}[h]
  \centering
  \includegraphics[width=\textwidth]{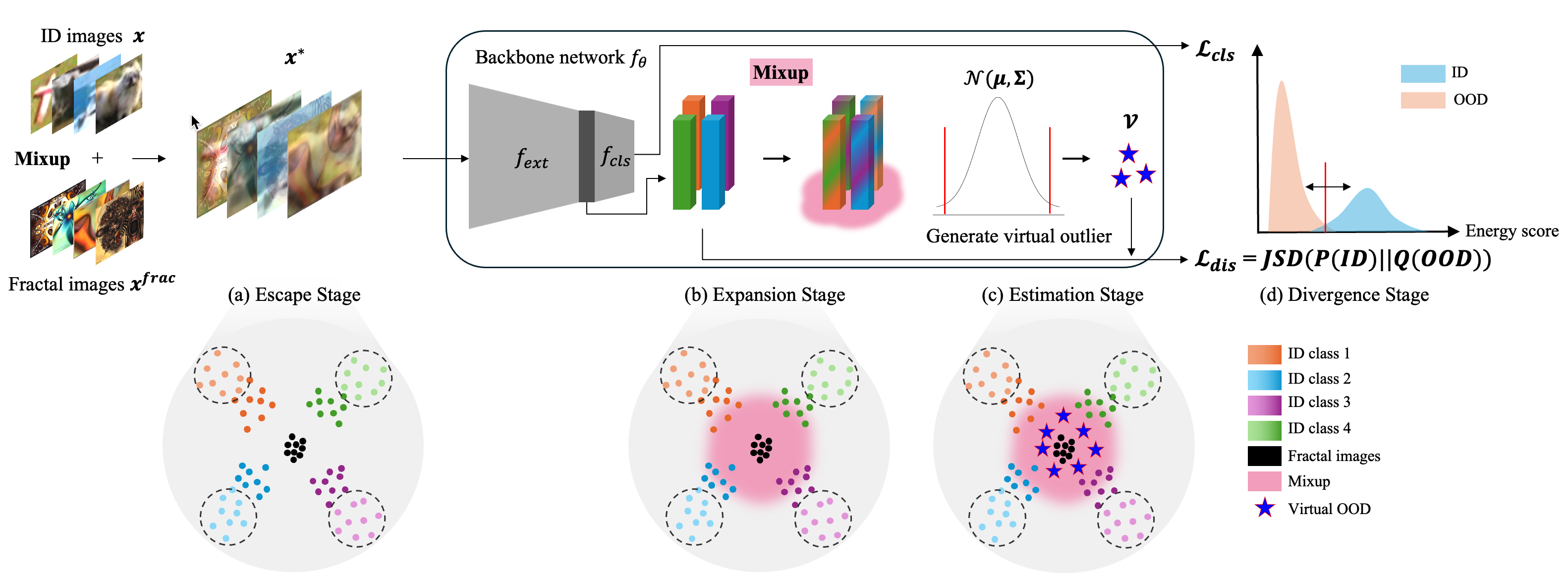}
  \caption{Overview process of ARES. (a) Escape stage generates surrogate ID instances, $\mathbf{x}^*$; out of the original ID region represented in dotted circles. (b) Expansion stage generates the $\tilde{\mathbf{x}}^*$ region by Mixup $\mathbf{x}^*$. (c) Estimation stage synthesizes a virtual outlier by selecting the instance with the t-th smallest likelihood in the $\tilde{\mathbf{x}}^*$ region. (d) In Divergence stage, $\mathbf{x}^*$ from Escape stage is used as ID; while $\mathbf{v}$ from Estimation stage is used as OOD. }
  \label{fig:Overview}
  \vskip -0.32cm
\end{figure*}

\section{Method} \label{sec:Method}
\subsection{Problem Setting}
In this paper, we solve out-of-distribution (OOD) detection task, where we are given a training dataset consisting of $N$ number of instances, which is represented as $\mathcal{D} = \{(\mathbf{x}_i, \mathbf{y}_i)\}^N_{i=1}$.
Compared to the training dataset, the test dataset may contain instances of unseen class labels as well as instances which share the same class labels with $\mathcal{D}$.
For clarity, we simply denote the instances with known classes as ID instances or inliers; while we denote the instances of unknown classes as OOD instances or outliers.
 
The goal of OOD detection task is training a model, which is a neural network $f_{\theta}$ parameterized by $\theta$; where $f_{\theta}$ can be further represented as $f_{\theta}(x)=f_{cls}(f_{ext}(x))$. Here, $f_{ext}$ is a feature extractor which extracts a $d$-dimensional feature embedding for $x$; and $f_{cls}$ is a classifier that maps the feature embedding into a label space. The output of $f_{ext}$ is also called as feature at the penultimate layer of $f_{\theta}$, and $f_{cls}$ consists of fully-connected layers.
Overall, $f_{\theta}$ should predict the correct classes for the inliers as well as predict outliers as "unknown" class rather than one of the ID classes.

\subsection{Motivation}
For an effective OOD detection, we focus on the proposal of a sampling-based method.
The basic assumption for our method is that we should sample virtual outliers at the farthest region as possible from the given ID region. Having said that, every stage of our proposed method commonly aims to depart the original ID region, contributing to enhancing the values of generated virtual outliers.

\subsection{ARES: Auxiliary Range Expansion for Outlier Synthesis}
In this paper, we propose a novel approach named \underline{A}uxiliary \underline{R}ange \underline{E}xpansion for Outlier \underline{S}ynthesis, or ARES, whose goal is to generate virtual OOD instances at the as farthest as possible region from ID instances. For the sake of this goal, ARES consists of four stages to generate OOD instances, which we name as 1) Escape, 2) Expansion, 3) Estimation, and 4) Divergence stages.

\subsubsection{\textbf{Escape Stage}}
During the training phase, we are only allowed an access to the ID dataset which is given as a training set, by the natural assumption of OOD detection. The given ID instances become the only ingredient to generate OOD instances.
Hence, ARES starts with the Escape stage, which generates a surrogate ID set, denoted as $\mathcal{D}^*$, apart from the original ID dataset, denoteod as $\mathcal{D}$.
$\mathcal{D}^*$ actually plays the role as the starting point for OOD generation.

For an effective generation of $\mathcal{D}^*$, we adopt PixMix \cite{PixMix}, which mixes the original data with fractal images, denoted as $\mathcal{F}=\{(x_{i}^{frac})\}_{i=1}^L$, as the below.
\begin{gather}
    \label{eq:escape}
    x_i^* = \lambda_1 x_i + (1-\lambda_1) x_i^{frac}, \lambda_1 \sim Beta(\alpha_1,\alpha_1) \\
    \mathcal{D}^* = \{(\mathbf{x}_i^{*}, y_i)\}^N_{i=1} 
\end{gather}
The generated $\mathcal{D}^{*}$ keeps the original label, $y_i$, of the original instance, $x_i$.
Following prior work of PixMix, we apply either geometric transformation (e.g., rotation, solarization, etc.) or Mixup with a fractal image to each ID instance for up to four iterations. In this paper, we further restrict to choose Mixup with a fractal image for at least one iteration.

PixMix has been proven to be effective in OOD detection by integrating diverse natural patterns from fractal images. Moreover, fractal images used in the Escape stage of ARES are class-agnostic and relatively easy to get, compared to previous work that requires a heavy real OOD dataset \cite{MixOE}.
We have figured out that the generated $\mathcal{D}^{*}$ reflects the original characteristics of $\mathcal{D}$ but is located outside of the original ID region. That is, we have achieved the surrogate ID instances by escaping from the real ID region, as shown in Figure \ref{fig:Overview} (a).

\subsubsection{\textbf{Expansion Stage}}
The generated surrogate ID instances are far away from the original ID region, so we can now start the actual process of generating OOD instances.
To fully use the generated surrogate ID region $\mathcal{D}^{*}$, we expand this region by mixing multiple surrogate ID instances with each other, so that the expanded region could simulate the OOD region.
For the Expansion stage, we adopt Manifold Mixup \cite{Manifold_Mixup}, which mixes feature representations rather than raw data instances. In specific, we select the feature map at the penultimate layer of the classifier for mixing as the below:
\vskip -0.4cm
\begin{equation}
    \tilde{\mathbf{x}}^* = \lambda_2 f _{ext}(\mathbf{x}^*_i) + (1 - \lambda_2)f_{ext}(\mathbf{x}^*_j), \label{eq:expansion}
\end{equation}
where  $\lambda_2 \sim Beta(\alpha_2,\alpha_2)$ and $\mathbf{x}^*_i, \mathbf{x}^*_j \in \mathcal{D}^{*}$.
This allows us an efficient mixing process in a low-dimensional space as well as the utilization of rich information extracted by the feature extractor, $f_{ext}$.
We can assume this expanded region, which is represented as pink in Figure \ref{fig:Overview} (b), would contain OOD region which is far away from our given ID region.

Specifically, we need a sufficiently discriminative feature to get $\tilde{x}^*$ in Eq. \eqref{eq:expansion}. Hence, we define a pre-defined epoch to pre-train the feature extractor, $f_{ext}$, to start Eq. \eqref{eq:expansion}.


\subsubsection{\textbf{Estimation Stage}}
Finally, we get the expanded set, denoted as $\tilde{\mathcal{X}} = \{\tilde{\mathbf{x}}^*\}$. Given $\tilde{\mathcal{X}}$, we assume a multivariate Gaussian distribution of $\tilde{\mathcal{X}}$ as $\mathcal{N}(\boldsymbol\mu, \*{\Sigma})$, where we estimate the mean and variance of the distribution as the below.
\begin{align}
    \widehat{\mu}&=\frac{1}{N} \sum_{i=1}^N \tilde{\mathbf{x}}^*_i \label{eq:mu} \\ 
    \widehat{\mathbf{\Sigma}}&=\frac{1}{N} \sum_{i=1}^N\left(\tilde{\mathbf{x}}^*_i-\widehat{\mu}\right)\left(\tilde{\mathbf{x}}^*_i-\widehat{\mu}\right)^{\top} \label{eq:var}
\end{align}
It should be noted that the distribution on $\tilde{\mathcal{X}}$ is estimated as a class-agnostic form rather than a class-wise form, by using all $\tilde{\mathbf{x}}^*\in\tilde{\mathcal{X}}$ at once for calculating Eq. \eqref{eq:mu}--\eqref{eq:var}.
Modeling class-agnostic distribution further fits with the nature of OOD instances, which would be scattered across the whole data space.

Following the prior work \cite{du2021vos}, we assume that OOD instance would have the lowest likelihood. Hence, we generate virtual outliers as below:
\begin{align}
  & \mathcal{V}= \{ \mathbf{v} \vert \frac{1}{(2 \pi)^{p / 2}|\widehat{{{\Sigma}}}|^{1 / 2}} \exp \left(-\frac{1}{2}(\mathbf{v}-\widehat{\mu})^{\top} \widehat{{\Sigma}}^{-1}(\mathbf{v}-\widehat{\mu})\right) < \epsilon\},
   \label{eq:virtual}
\end{align}
where $p$ denotes the dimension of $\widehat{\mu}$, and $\epsilon$ is a hyperparameter for thresholding the likelihood of outliers.
To set the value of $\epsilon$, we randomly select $M$ instances from $\tilde{\mathcal{X}}$; and we use the $t$-th smallest likelihood as $\epsilon$. We describe the selection on $M$ and $t$ in Section \ref{exp:config}.
Figure \ref{fig:Overview} (c) confirms that the generated virtual outliers, $\mathbf{v}$, which are denoted as blue stars, are located far away from the original ID region.
The sampled $\mathbf{v}$ are generated from broad region, which is modeled via Escape and Expansion stages. That is, $\mathbf{v}$ covers from fine-grained (i.e., $\tilde{\mathcal{X}}$) to coarse-grained (i.e., $\mathcal{F}$) OOD, enhancing the robustness of ARES.

\subsubsection{\textbf{Divergence Stage}}
Given the generated virtual outliers, we need to train a discriminator that distinguishes virtual outliers from the inliers that were given as the training set.
Traditionally, an energy-based score has been widely used because of its simplicity.

In prior work \cite{du2021vos}, the discriminator on ID and OOD uses cross-entropy loss for energy-based score with further forwarding them through logistic regression.
However, we conjecture that this loss modeling has limitations in that logistic regression requires additional parameters to train. Hence, we propose to directly discriminate energy-based scores between inliers and virtual outliers, by adopting Jensen-Shannon Divergence (JSD) loss \cite{JSD}. JSD is widely used as loss in generative models like GAN when learning the difference between two distributions, because it allows for calculating the distance as divergence \cite{GAN_Ian}.
JSD often invokes problems when used as a minimization objective in generative models such as Wasserstein Generative Adversarial Networks (WGAN), because of its inaccurate expression of the distance metric when applied to already distinguished distributions \cite{WGAN}.
However, these concerns are alleviated in ARES where JSD is used as a maximization objective to separate overlapping distributions.

Hence, we adopt JSD loss for ARES, which attempts to directly repel the energy-based scores between inliers and outliers. In specific, since $x^*$ are in raw pixel space while $\mathbf{v}$ are in feature space extracted by $f_{ext}$, we first extract features for $x^*$ using $f_{ext}$. Then, we train our OOD discriminator with loss as below:
\begin{equation}
    \mathcal{L}_\text{dis}=\mathbb{E}_{\mathbf{v}\sim \mathcal{V}, \mathbf{x}^* \sim \mathcal{D^*}} \big[ JSD \left( {\mathcal P} \left( E(f_{ext}(\mathbf{x}^*);\theta) \right) \parallel {\mathcal Q} \left( E(\mathbf{v};\theta) \right) \right) \big], \label{eq:jsdloss}
\end{equation}
where $\mathcal P$ and $\mathcal Q$ denote the probability distributions of energy-based scores of ID instances and virtual OOD instances, respectively.

To construct the probability distribution of energy-based score, we sample virtual OOD instances using Eq. \eqref{eq:virtual} with the same number as ID instances.
That is, given ID instances with batch size of $B$, $\epsilon$ in Eq. \eqref{eq:virtual} is set as the $B$-th lowest likelihood value.
With the sampled virtual OOD instances and ID instances, we construct a frequency distribution of energy-based scores for both instances.
Then, we assume Gaussian distributions for inliers and virtual outliers, denoted as $\mathcal{P} \sim \mathcal{N}(\boldsymbol\mu_\mathbf{x^*}, \mathbf{\sigma}_{\mathbf{x^*}}^2)$ and $\mathcal{Q} \sim \mathcal{N}(\boldsymbol\mu_\mathbf{v}, \mathbf{\sigma}_{\mathbf{v}}^2)$ respectively, by estimating the mean and variance as followings.

\begin{align}
    & \begin{cases}
        \mu_\mathbf{x^*}=\frac{1}{B} \sum_{i=1}^B E(f_{ext}(\mathbf{x}^*_i);\theta) \\
        \mathbf{\sigma}_\mathbf{x^*}=\frac{1}{B} \sum_{i=1}^B\left(E(f_{ext}(\mathbf{x}^*_i);\theta)-\mu_\mathbf{x^*}\right)^2 \label{eq:mu_jsd}
    \end{cases} \\[0.5em]
    & \begin{cases}
        \mu_\mathbf{v}=\frac{1}{B} \sum_{i=1}^B E(\mathbf{v}_i;\theta) \\
        \mathbf{\sigma}_\mathbf{v}=\frac{1}{B} \sum_{i=1}^B\left(E(\mathbf{v}_i;\theta)-\mu_\mathbf{v}\right)^2 \label{eq:sigma_jsd}
    \end{cases}
\end{align}

Finally, the JSD term in the discrimination loss of Eq. \eqref{eq:jsdloss} is calculated using the estimated Gaussian distributions:
\begin{align}
    JSD \left( {\mathcal P} \parallel {\mathcal Q} \right) = \frac{1}{2} \big[ KLD \left( {\mathcal P} \parallel \mathcal{M} \right) + KLD \left( {\mathcal Q} \parallel \mathcal{M} \right) \big], \label{eq:jsd_kld}
\end{align}
where $\mathcal{M}=\frac{\mathcal{P}+\mathcal{Q}}{2}$ follows $\mathcal{N}(\boldsymbol\mu_M, \mathbf{\sigma}_M^2)$ and each $KLD$ term is calculated in a closed-form solution as the followings.
\begin{align}
    \begin{cases}
        KLD \left( {\mathcal P} \parallel {\mathcal M} \right) = log\frac{\sigma_\mathbf{M}}{\sigma_{\mathbf{x}^*}}+\frac{\sigma_{\mathbf{x}^*}^2+(\mu_{\mathbf{x}^*}-\mu_\mathbf{M})^2}{2\sigma_\mathbf{M}^2}-\frac{1}{2} \\
        KLD \left( {\mathcal Q} \parallel {\mathcal M} \right) = log\frac{\sigma_\mathbf{M}}{\sigma_{\mathbf{v}}}+\frac{\sigma_{\mathbf{v}}^2+(\mu_{\mathbf{v}}-\mu_\mathbf{M})^2}{2\sigma_\mathbf{M}^2}-\frac{1}{2} \label{eq:kld_m}
    \end{cases}
\end{align}

In Eq. \eqref{eq:jsdloss}--\eqref{eq:kld_m}, $E(\mathbf{x};\theta)$ denotes the energy-based score of an arbitrary input, $\mathbf{x}$, as the below:
\begin{equation}
E(\mathbf{x};\theta) = -\log \sum_{k=1}^K w_k\cdot \exp({f_{cls}(\mathbf{x})}), \label{eq:energy}
\end{equation}
where $k$ denotes the dimension of $f_{cls}(\mathbf{x})$ and $w_k$ is a learnable weight coefficient.

\subsection{Overall Algorithm}
The model, $f_{\theta}$, is trained to minimize a total loss as the below:
\begin{equation}
    \mathcal{L}_\text{total} = \mathbb{E}_{(\mathbf{x}^*, {y}) \sim \mathcal{D^*}}\left[\mathcal{L}_\text{cls}\right]+\beta \cdot \frac {1}{\mathcal{L}_\text{dis}}, \label{eq:finaljsdloss}
\end{equation}
which aims at both classifying inliers correctly by $\mathcal{L}_\text{cls}$; as well as discriminating between inliers and outliers by $\mathcal{L}_\text{dis}$.
Here, $\beta$ controls the weight for the discrimination loss of Eq. \eqref{eq:jsdloss}.

After training the model, we discriminate the test dataset into inlier or outlier using the energy-based score as a criterion, which follows Eq. \eqref{eq:discrimination} of the below:
\begin{equation}
    G(\mathbf{x}_{test})=\left\{\begin{array}{ll}
1 & \text { if }E(\mathbf{x}_{test}; \theta)\geq \gamma, \\
0 & \text { if }E(\mathbf{x}_{test}; \theta) <\gamma,
\end{array}\right.\label{eq:discrimination}
\end{equation}
where 1 denotes the inlier and 0 denotes the outlier.
If a test example is discriminated as an inlier, we predict its label using the prediction of $f_{\theta}$. Otherwise, we predict its label as "unknown".
In Eq. \eqref{eq:discrimination}, $\gamma$ is a threshold that is used to separate ID instances from OOD instances. In this paper, we set $\gamma$ to screen ID instances at FPR95. 

Finally, Algorithm \ref{alg:ARES} describes the overall process of ARES.
\begin{algorithm}
\caption{ARES: Auxiliary Range Expansion for outlier Synthesis}
\label{alg:ARES}
\begin{algorithmic}[1] 
\Require ID data \(\mathcal{D} = \{(\mathbf{x}_i, y_i)\}^N_{i=1}\), 
  \Statex \hspace{\algorithmicindent}Fractal data \(\mathcal{F} = \{(\mathbf{x}_i^{frac})\}^L_{i=1}\)
\Ensure OOD detection model \(f_{\theta}\)
  \Statex \textcolor{gray}{\LeftComment{Escape stage}}
  \State Load the original ID data, \(\mathcal{D}\), and auxiliary fractal data, \(\mathcal{F}\)
  \State Get $\mathcal{D}^* = \{(\mathbf{x}_i^{*}, y_i)\}^N_{i=1}$ by mixing ID data and fractal data as Eq. \eqref{eq:escape}
\While{train}
  \Statex \hspace{\algorithmicindent} \textcolor{gray}{\LeftComment{Expansion stage}}
  \State Extract feature embeddings for $\mathbf{x}^* \in \mathcal{D}^*$
  \State Mixup $f_{ext}(\mathbf{x}^*)$ with each other as Eq. \eqref{eq:expansion}
  \State Construct $\tilde{\mathcal{X}}$ with mixed $f_{ext}(\mathbf{x}^*)$
  \Statex \hspace{\algorithmicindent} \textcolor{gray}{\LeftComment{Estimation stage}}
  \State Estimate the multivariate Gaussian distribution for $\tilde{\mathcal{X}}$
  \Statex \hspace{\algorithmicindent}as Eq. \eqref{eq:mu}--\eqref{eq:var}
  \State Sample virtual outliers $\mathbf{v}$ using Eq. \eqref{eq:virtual}
  \Statex \hspace{\algorithmicindent} \textcolor{gray}{\LeftComment{Divergence stage}}
  \State Calculate the JSD discrimination loss using energy-
  \Statex \hspace{\algorithmicindent}based score and Eq. \eqref{eq:jsdloss} 
  \State Update the OOD detection model, \(f_{\theta}\), using Eq. \eqref{eq:finaljsdloss}
\EndWhile
\While{eval}
  \State Calculate Eq. \eqref{eq:energy} for test dataset
  \State Perform thresholding comparison using Eq. \eqref{eq:discrimination}
\EndWhile
\end{algorithmic}
\end{algorithm}

\begin{table*}[h]
  \caption{OOD detection performance of ARES and baselines on CIFAR-10 and CIFAR-100 as ID. Values are percentages. The best performance is represented as bold, and the second-best performance is represented as underline. $\uparrow$ indicates that larger values are better, and $\downarrow$ indicates that smaller values are better. The results are replicated three times.}
  \label{tab:main_results}
  \begin{subtable}{\textwidth}
    \caption{Performance on CIFAR-10 as ID}
    \label{tab:cifar10}
    \resizebox{\textwidth}{!}{
  \begin{tabular}{@{}lccccccccccccc@{}}
    \toprule
    & \multicolumn{12}{c}{\textbf{OOD Datasets}} \\
    \cmidrule{2-13}
    \textbf{Methods}
    & \multicolumn{2}{c}{\textbf{Texture}} 
    & \multicolumn{2}{c}{\textbf{SVHN}} 
    & \multicolumn{2}{c}{\textbf{LSUN-C}}
    & \multicolumn{2}{c}{\textbf{iSUN}}
    & \multicolumn{2}{c}{\textbf{Place365}}
    & \multicolumn{2}{c}{\textbf{Average}} \\
    \cmidrule{2-13}
     & \textbf{FPR95$\downarrow$} & \textbf{AUROC$\uparrow$}
     & \textbf{FPR95$\downarrow$} & \textbf{AUROC$\uparrow$}
     & \textbf{FPR95$\downarrow$} & \textbf{AUROC$\uparrow$}
     & \textbf{FPR95$\downarrow$} & \textbf{AUROC$\uparrow$}
     & \textbf{FPR95$\downarrow$} & \textbf{AUROC$\uparrow$}
     & \textbf{FPR95$\downarrow$} & \textbf{AUROC$\uparrow$} \\
    \midrule
    MSP \cite{MSP} & 66.45 & 88.50 & 59.66 & 91.25 & 45.21 & 93.80
                       & 54.57 & 92.12 & 62.46 & 88.64 & 57.67 & 90.86 \\
    ODIN \cite{ODIN} & 68.63 & 86.92 & 67.64 & 84.49 & 22.62 & 95.33 
                          & 30.89 & 94.09 & 71.06 & 84.93 & 52.17 & 89.15 \\
    Mahalanobis \cite{Mahalanobis} & 33.51 & 92.00 & 34.57 & 88.80 & 75.99 & 62.53 
                          & 43.56 & 86.79 & 61.47 & 79.84 & 49.82 & 81.99 \\
    CSI \cite{CSI} & 57.89 & 82.38 & 32.88 & 89.04 & 21.70 & 93.37 
                          & 29.76 & 91.52 & 45.73 & 84.77 & 37.59 & 88.22 \\
    MixOE \cite{MixOE} & 34.59 & 91.54 & 16.58 & 94.95 & 24.23 & 93.04 
                          & 48.78 & 88.69 & 36.90 & 91.42 & 32.22 & 91.93 \\
    VOS \cite{du2021vos} & 45.98 & 87.80 & 35.92 & 92.84 & 8.38 & \underline{98.25} 
                          & 29.53 & 93.70 & 41.53 & 89.59 & 32.27 & 92.44 \\
    RONF \cite{RONF} & - & - & - & - & - & - & -
                          & - & - & - & 22.61 & 95.34 \\
    NPOS \cite{NPOS} & 21.57 & 93.14 & \underline{9.23} & \underline{98.45} & 12.90 & 94.87 
                          & 21.43 & 92.36 & 30.89 & 90.18 & 19.20 & 93.80 \\
    Dream-OOD \cite{Dream} & \underline{12.32} & \underline{93.21} & 10.43 & 94.32 & \underline{7.76} &                                         97.82 & \underline{9.11} & \underline{94.17} & \textbf{5.32} & \textbf{98.56} & \underline{8.99} & \underline{95.42} \\
    \textbf{ARES} (ours) & \textbf{4.32} & \textbf{99.01} & \textbf{4.98} & \textbf{99.04} & \textbf{6.12} 
                         & \textbf{98.70} & \textbf{4.75} & \textbf{98.94} & \underline{22.82} & \underline{94.81} & \textbf{8.60} & \textbf{98.10} \\
    \bottomrule
  \end{tabular}
  }
  \end{subtable}
  \vskip 0.4cm
  \begin{subtable}{\textwidth}
    \caption{Performance on CIFAR-100 as ID}
    \label{tab:cifar100}
    \resizebox{\textwidth}{!}{
  \begin{tabular}{@{}lcccccccccccc@{}}
    \toprule
    & \multicolumn{12}{c}{\textbf{OOD Datasets}} \\
    \cmidrule{2-13}
    \textbf{Methods}
    & \multicolumn{2}{c}{\textbf{Texture}} 
    & \multicolumn{2}{c}{\textbf{SVHN}} 
    & \multicolumn{2}{c}{\textbf{LSUN-C}}
    & \multicolumn{2}{c}{\textbf{iSUN}}
    & \multicolumn{2}{c}{\textbf{Place365}}
    & \multicolumn{2}{c}{\textbf{Average}} \\
    \cmidrule{2-13}
     & \textbf{FPR95$\downarrow$} & \textbf{AUROC$\uparrow$}
     & \textbf{FPR95$\downarrow$} & \textbf{AUROC$\uparrow$}
     & \textbf{FPR95$\downarrow$} & \textbf{AUROC$\uparrow$}
     & \textbf{FPR95$\downarrow$} & \textbf{AUROC$\uparrow$}
     & \textbf{FPR95$\downarrow$} & \textbf{AUROC$\uparrow$}
     & \textbf{FPR95$\downarrow$} & \textbf{AUROC$\uparrow$} \\
    \midrule
    MSP \cite{MSP} & 86.45 & 71.32 & 85.30 & 72.41 & 85.55 & 74.00
                       & 88.55 & 68.59 & 73.40 & 81.09 & 83.85 & 73.48 \\
    ODIN \cite{ODIN} & 85.75 & 73.17 & 89.50 & 76.13 & 74.70 & 83.93
                         & 90.20 & 68.27 & \textbf{41.50} & \textbf{91.60} & 76.33 & 78.62 \\
    Mahalanobis \cite{Mahalanobis} & 77.54 & 71.09 & 80.27 & 60.30 & 89.84 & 52.14 
                          & 82.10 & 61.53 & 87.40 & 54.73 & 83.43 & 59.96 \\
    CSI \cite{CSI} & 91.99 & 54.61 & 59.80 & 80.62 & 50.63 & 84.86 
                          & 78.90 & 69.81 & 91.27 & 57.69 & 74.52 & 69.52 \\
    MixOE \cite{MixOE} & 68.35 & 76.59 & 75.01 & 69.88 & 47.72 & 83.28 
                          & 68.64 & 72.01 & \underline{56.40} & \underline{80.1} & 63.22 & 76.39 \\
    VOS \cite{du2021vos} & 81.77 & 76.73 & 81.83 & 81.64 & 37.12 & 93.68
                          & 77.12 & 76.02 & 80.37 & 76.11 & 71.64 & 80.84 \\
    RONF \cite{RONF} & - & - & - & - & -
                          & - & - & - & - & - & 45.49 & 86.89 \\
    NPOS \cite{NPOS} & \underline{42.59} & \underline{90.66} & \textbf{18.95} & \textbf{97.03} & 63.77 & 77.28 
                          & 51.22 & 83.43 & 69.07 & 73.67 & 49.12 & 84.41 \\
    Dream-OOD \cite{Dream} & 48.10 & 88.02 & 58.90 & 86.69 & \underline{28.85} & \underline{95.10}
                          & \textbf{1.15} & \textbf{99.72} & 74.85 & 78.41 & \underline{42.37} & \underline{89.58} \\
    \textbf{ARES} (ours) & \textbf{28.55} & \textbf{93.28} & \underline{33.40} & \underline{93.38} & \textbf{24.48} & \textbf{95.68} & \underline{38.90} & \underline{91.30} & 65.32 & 82.43 & \textbf{38.13} 
    & \textbf{91.21} \\
    \bottomrule
  \end{tabular}
  }
  \end{subtable}
\end{table*}

\section{Experiments} \label{sec:Experiment}
\subsection{Experimental Settings}
\subsubsection{\textbf{Dataset}}  We train our model on the ID dataset with CIFAR-10 and CIFAR-100 \cite{cifar10}. For evaluation of the OOD detection, we use five datasets that can cover various real-world out-of-distribution; including 1) Texture \cite{Textures}, 2) SVHN \cite{SVHN}, 3) LSUN-C \cite{LSUN}, 4) iSUN \cite{iSUN}, 5) Place365 \cite{place365}. Each dataset is exclusive to ID dataset. 

\subsubsection{\textbf{Model and Training Details}} \label{exp:config} We use WideResNet40 \cite{Wide_ResNet} for the backbone network. We adopt Stochastic Gradient Descent (SGD) optimizer \cite{SGD} with a learning rate of 1e-1 which decreases to 1e-6 using a cosine scheduler \cite{cosine_scheduler}.
We use a batch size of 128 for both ID instances and OOD instances to calculate the loss of Eq. \eqref{eq:finaljsdloss}. In specific, to sample virtual outliers, $\mathbf{v}$, we randomly selected $M=10,000$ instances and used the instances with the $t=128$-th smallest likelihood to set $\epsilon$ of Eq. \eqref{eq:virtual}.
We train our model for 500 epochs. During the first 200 epochs, the model is trained with only classification loss, which is the first term of Eq. \eqref{eq:finaljsdloss}. Then, the model is trained with both classification loss and discrimination loss for the rest epochs.
Additionally, detailed settings for ARES are as follows. In the Escape stage, the Mixup with the fractal data uses $\alpha_1$=3 in Eq. \eqref{eq:escape}. In the Expansion stage, we use $\alpha_2$=2 in Eq. \eqref{eq:expansion} for the Mixup in the feature space. Finally, $\beta$, which is the weight for the discrimination loss, in Eq. \eqref{eq:finaljsdloss} is 0.1.

\subsubsection{\textbf{Evaluation Metrics}} We use various metrics to evaluate the performance of OOD detection:
1) FPR95, which is the false positive rate of OOD samples when the true positive rate of ID samples is at 95\%, and 2) AUROC, which is the area under the receiver operating characteristic curve.  

\subsubsection{\textbf{Baseline Method}} As presented in Table \ref{tab:main_results}, we compare our ARES with other OOD detection methods. The methods compared include a total of 8, as follows:
1) MSP \cite{MSP}, 2) ODIN \cite{ODIN}, 3) Mahalanobis \cite{Mahalanobis}, 4) CSI \cite{CSI}, 5) MixOE \cite{MixOE}, 6) VOS \cite{du2021vos}, 7) RONF \cite{RONF} and 8) NPOS \cite{NPOS}.
We implemented most of the baselines through the Open-OOD benchmark framework \cite{Open-OOD}.
For MSP and ODIN, we adopted the result from relevant research \cite{NPOS}.
For RONF, we could not reproduce the model so we recorded the averaged results reported by the authors.

\subsection{Main Results}

\begin{figure}[t]
  \centering
  \begin{subfigure}{0.5\textwidth} 
    \includegraphics[width=\textwidth]{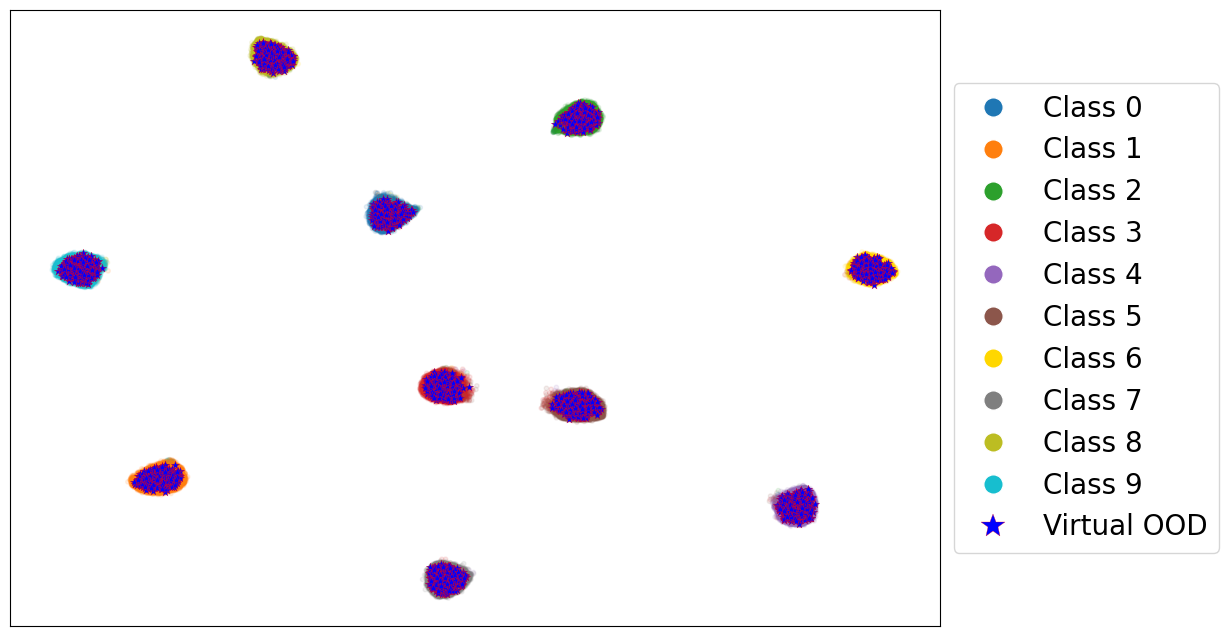}
    \caption{UMAP of feature embeddings from VOS}
    \label{fig:vos_umap}
  \end{subfigure}
  \vskip 0.4cm
  \begin{subfigure}{0.5\textwidth}
    \includegraphics[width=\textwidth]{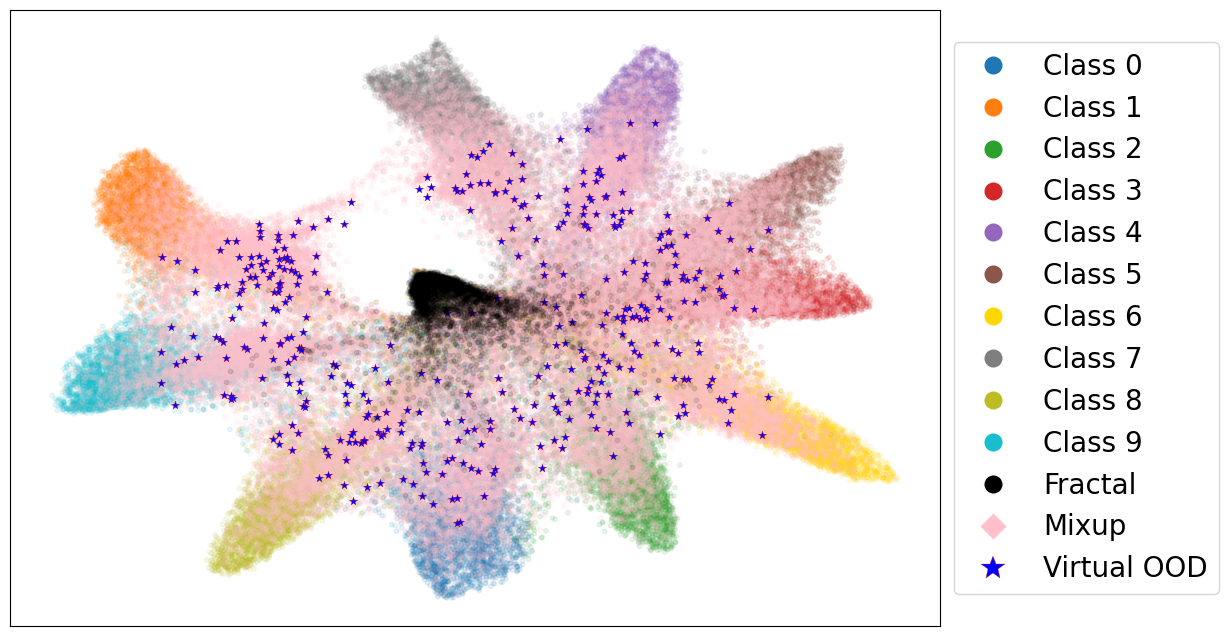}
    \caption{UMAP of feature embeddings from ARES}
    \label{fig:ares_umap}
  \end{subfigure}
  \caption{Comparison of UMAP on CIFAR-10 as ID}
  \label{fig:main_umap}
  \vskip -0.4cm
\end{figure}

\subsubsection{\textbf{Quantitative Analysis Results}} We first provide quantitative results on evaluation metrics in Table \ref{tab:main_results}. As shown in Table \ref{tab:cifar10}, our methodology, ARES, demonstrates the best performance in all metrics across various OOD datasets when using CIFAR-10 as ID dataset.

If we contrast ARES with VOS \cite{du2021vos} which is our most relevant baseline, the gap is apparent. The largest gap is seen on the Texture dataset, where FPR95 and AUROC improved from 45.98\% to 4.32\%, and from 87.80\% to 99.01\%, respectively. Based on the average values, improvements are obvious from 32.27\% to 8.60\% in FPR95, and from 92.44\% to 98.10\% in AUROC.
The primary reason for these significant improvements lies in the process of estimating distributions of OOD region to synthesize virtual outliers.
VOS estimates the region for virtual OOD by only utilizing given ID instances. This leads to a discernible value difference between the estimated virtual outliers and real outliers.
However, ARES aims to model the OOD region as far from ID region as possible, which is accomplished by each stage of Escape, Expansion, Estimation, and Divergence stages.
Hence, ARES generates more OOD-like virtual instances that deviate from original ID instances.

Additionally, we compare ARES to RONF \cite{RONF}, which has the second-best performance of AUROC. With regard to AUROC, ARES with 98.10\% outperforms RONF with 95.34\%, which shows an improvement of 2.76 percentage point.
RONF mixes virtual ID instances found at the boundary to generate virtual OOD instances. That is, the ingredient of Mixup lies near the ID region.
Also, RONF uses the Mixup instances directly as virtual outliers.
Moreover, RONF still generates virtual OOD instances conditioned on class labels, which might limit the dispersion of OOD instances.
Compared to RONF, ARES adopts the Escape stage to find the region deviated from ID space for Mixup.
Also, ARES further estimates the distribution of OOD from the Mixup instances via Expansion stage.
Finally, ARES generates virtual outliers in a class-agnostic way, which reflects the scattered nature of real OOD instances.

When compared to NPOS \cite{NPOS}, which has the second-best performance of FPR95 based on average values, ARES improved from 19.20\% to 8.60\%, showing an enhancement of 10.60 percentage point.
NPOS proposes a non-parametric method to generate virtual outliers. Specifically, NPOS finds inliers lying at the boundary by utilizing k-NN distance and generates virtual outliers near the discovered inliers. Though NPOS differs from RONF in that it adopts a non-parametric method, NPOS still shares a similarity with RONF by limiting the generated virtual outliers near the ID region. Hence, ARES also outperforms NPOS by escaping from the ID region for the generation of OOD instances.

In Table \ref{tab:cifar100}, it can be observed that even when the ID dataset is CIFAR-100, our method demonstrates the best performance across all metrics on Texture and iSUN as OOD. Also, our method shows the second-best performance on SVHN and LSUN-C.
It should be noted that the average values across all OOD datasets are the best in ARES.

Especially, the improvement of ARES compared to other baselines was larger on CIFAR-100 than on CIFAR-10. For example, comparing ARES and CSI on Texture dataset, the improvement of FPR95 by ARES was 63.44 percentage point (i.e., from 91.99\% to 28.55\%) when using CIFAR-100 as ID dataset; while 53.57 percentage point (i.e., from 57.89\% to 4.32\%) were improved by ARES when using CIFAR-10 as ID dataset. This improvement by ARES is also observed on other datasets and baselines, suggesting that the effect of ARES becomes apparent in the more complex dataset.
 
\subsubsection{\textbf{Qualitative Analysis Results}} 
Now, we provide a qualitative analysis of the improvement by ARES.
Figure \ref{fig:main_umap} describes the distribution of actual inliers as well as virtual outliers using Uniform Manifold Approximation and Projection (UMAP) \cite{UMAP}.
In the Figure, the colored dots indicate the inliers, while blue stars indicate the virtual outliers.
As shown in Figure \ref{fig:vos_umap}, VOS synthesizes virtual outliers at the boundaries and further within each region of ID class.
The overlap of virtual outliers with inliers may diminish their value as outliers.
Conversely, ARES successfully synthesizes virtual outliers outside the ID class region, as shown in Figure \ref{fig:ares_umap}. We conjecture that the separation between inliers and outliers comes from each stage of ARES. 
As a result, ARES ensures the definitive value of virtual outliers, leading to an effective generation of OOD-like instances.

\begin{figure}[t]
  \centering
  \begin{subfigure}{0.235\textwidth}
    \includegraphics[width=\textwidth]{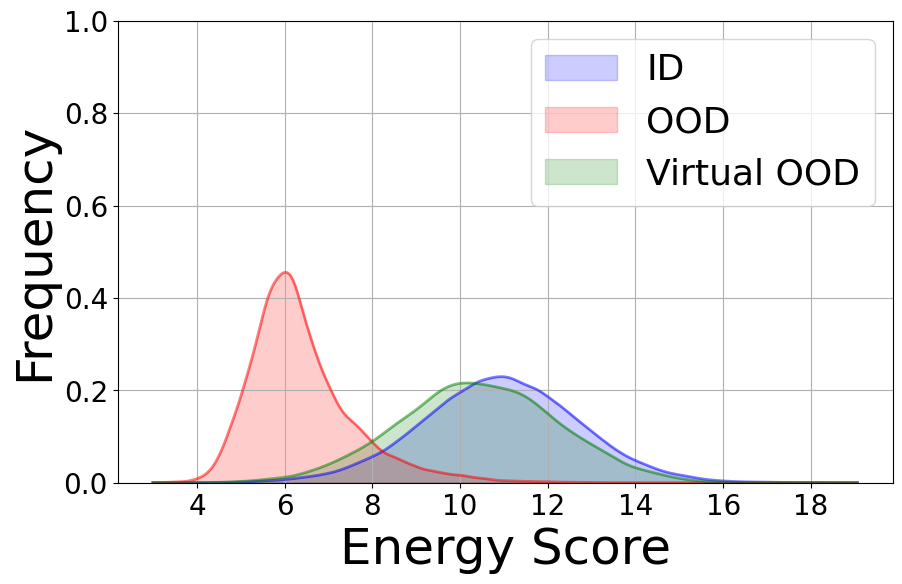}
    \captionsetup{justification=centering}
    \caption{Energy-based score \\ of VOS}
    \label{fig:VOS_final_energy}
  \end{subfigure}
  \hfill
  \begin{subfigure}{0.235\textwidth}
    \includegraphics[width=\textwidth]{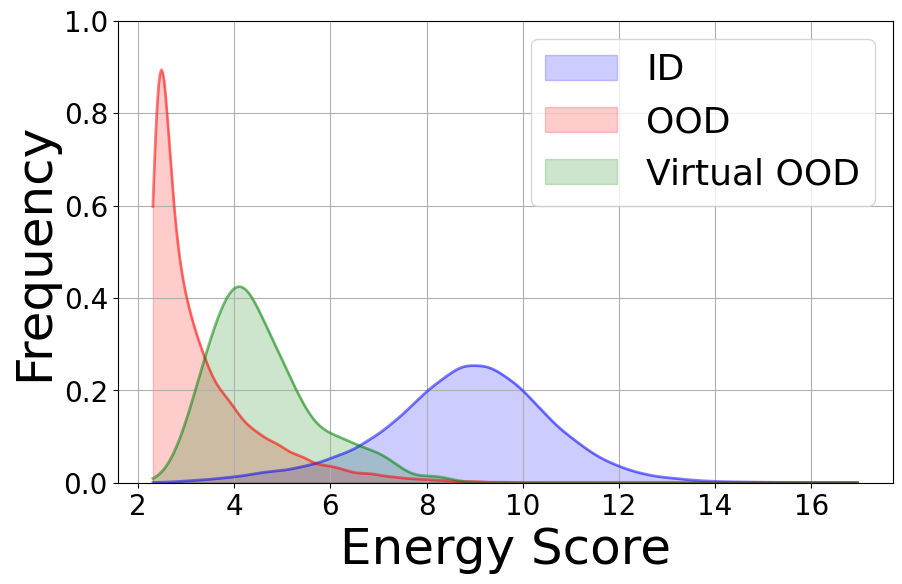}
    \captionsetup{justification=centering}
    \caption{Energy-based score \\ of ARES}
    \label{fig:ARES_final_energy}
  \end{subfigure}
  \caption{Comparison of Energy-based score distribution on CIFAR-10 as ID and SVHN as OOD}
  \label{fig:energy-pdf}
  \vskip -0.4cm
\end{figure}

Next, we describe the distribution of energy-based scores for inliers, outliers, and virtual outliers. In Figure \ref{fig:energy-pdf}, we compare VOS and ARES by selecting CIFAR-10 as ID and SVHN as OOD datasets. Figure \ref{fig:VOS_final_energy} shows that in VOS, energy-based scores of virtual outliers are distributed by overlapping with inliers rather than outliers, indicating the limited characteristics of virtual outliers as OOD instances. This further results in insufficient discrimination between inliers and outliers.
In contrast, Figure \ref{fig:ARES_final_energy} shows that virtual outliers by ARES have a closer distribution of energy-based scores to outliers rather than inliers. Having said that, ARES successfully discriminates outliers from inliers.

\begin{figure}[t]
  \centering
  \begin{subfigure}{0.46\textwidth} 
    \includegraphics[width=\textwidth]{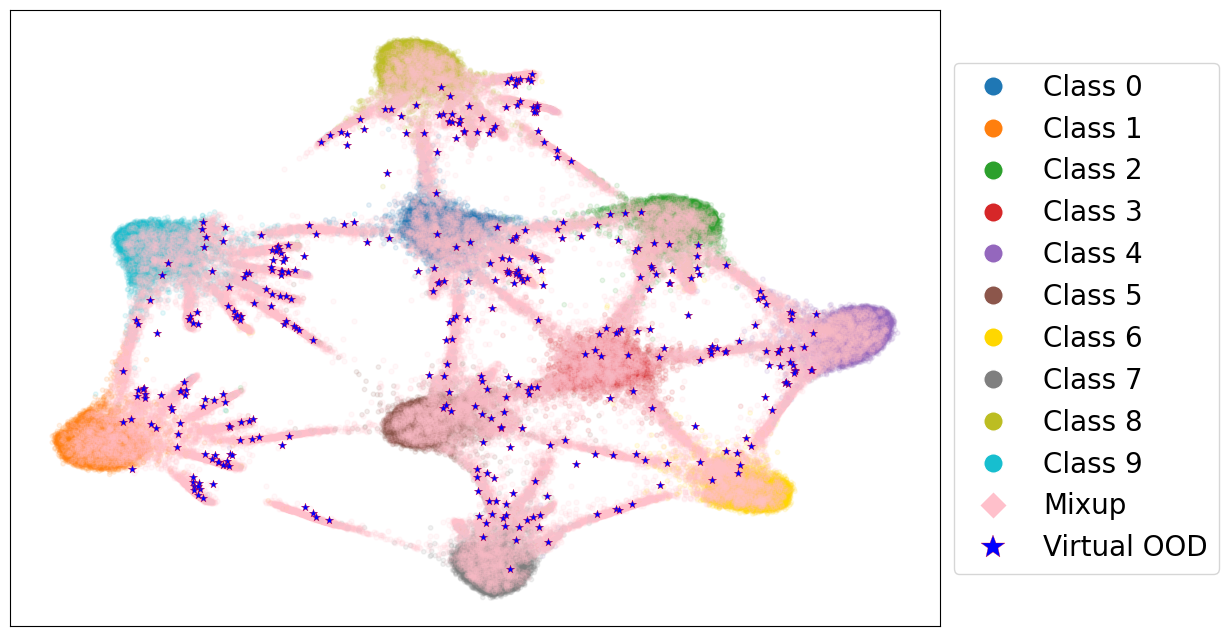}
    \caption{ARES w/o Escape}
    \label{fig:ares_wo_escape}
  \end{subfigure}
  \vskip 0.4cm
  \begin{subfigure}{0.46\textwidth}
    \includegraphics[width=\textwidth]{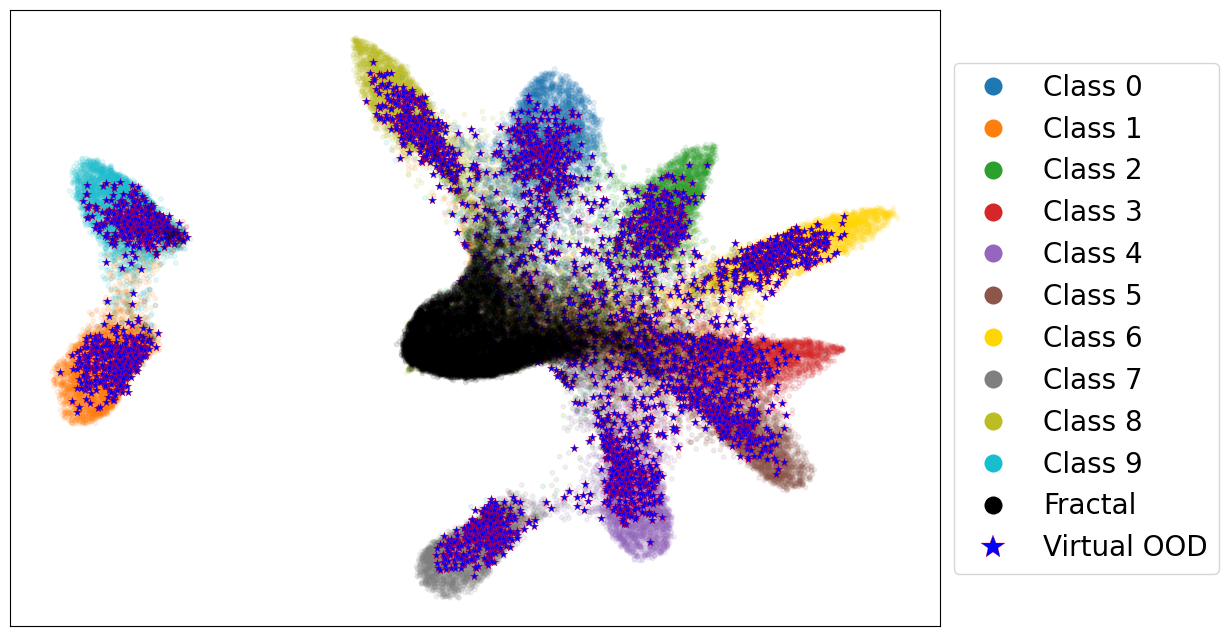}
    \caption{ARES w/o Expansion}
    \label{fig:ares_wo_expansion}
  \end{subfigure}
  \vskip 0.4cm
  \begin{subfigure}{0.46\textwidth}
    \includegraphics[width=\textwidth]{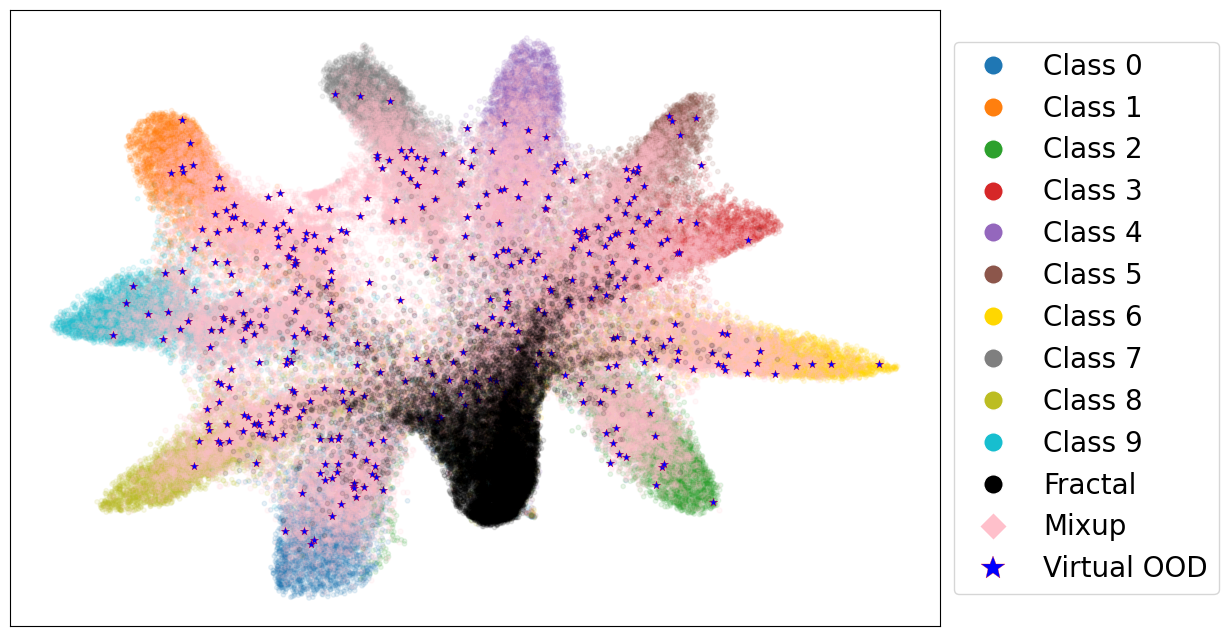}
    \caption{ARES w/o Estimation}
    \label{fig:ares_wo_estimation}
  \end{subfigure}
  \caption{Comparison of UMAP variations on CIFAR-10 as ID, by excluding each stage}
  \label{fig:ares_stageabl}
  \vskip -0.5cm
\end{figure}

\subsection{Ablation Study Results}
ARES is built upon the four main stages proposed in Section \ref{sec:Method}. Among them, the initial three stages, which are the Escape, Expansion, and Estimation stages, ultimately determine the distribution and shape of the generated virtual outliers. Then, Divergence stage serves as a discrimination loss. Hence, we provide a broad ablation study on each stage.

\begin{table}[t]
  \caption{FPR95$\downarrow$ of stage ablation studies on CIFAR-10 as ID and LSUN-C as OOD. Values are percentages. The best performance is represented as bold.}
  \label{tab:ablationstage}
  \resizebox{\columnwidth}{!}{
  \begin{tabular}{ccccc}
    \toprule
    \multirow{2}{*}{\textbf{Methods}} & \multicolumn{3}{c}{\textbf{Stages}} & \multirow{2}{*}{\textbf{LSUN-C}} \\
    \cmidrule(lr){2-4}
    & \textbf{Escape} & \textbf{Expansion} & \textbf{Estimation} & \\
    \midrule
    ARES w/o Escape & - & \checkmark & \checkmark & 11.25 \\
    \midrule
    ARES w/o Expansion & \checkmark & - & \checkmark & 7.05 \\
    \midrule
    ARES w/o Estimation & \checkmark & \checkmark & - & 6.75 \\
    \midrule
    ARES (w/ All stage) & \checkmark & \checkmark & \checkmark & \textbf{6.00} \\
    \bottomrule
  \end{tabular}
  }
  \vskip -0.5cm
\end{table}

\subsubsection{\textbf{Ablation on Each Stage of ARES}} 
Figure \ref{fig:ares_stageabl} shows the results by excluding each stage from ARES. In the Figure, colored dots denote inliers; black dots denote fractal images used in Escape stage; pink diamonds denote Mixup instances generated in Expansion stage; and blue stars denote virtual outliers generated in Estimation stage.

In Figure \ref{fig:ares_wo_escape}, which excludes the first Escape stage, we observe that the virtual outliers are synthesized near the ID region without significantly deviating from the ID boundary, which would have limited value of synthesized outliers to serve as an OOD instance.
We conjecture that a similar phenomenon would be observed in RONF, leading to a lower performance as in Table \ref{tab:cifar10}; because RONF first finds boundary instances following VOS and then further mixes them for virtual outliers.
Next, we exclude the Expansion stage in Figure \ref{fig:ares_wo_expansion}, where Mixup instances do not exist. In the figure, the virtual outliers are synthesized from the escaped ID region but are not expanded, resulting in a less broad distribution.
Finally, Figure \ref{fig:ares_wo_estimation} shows the result of excluding the Estimation stage and randomly selecting virtual outliers. In the figure, the UMAP is similar to Figure \ref{fig:ares_umap} which applies all the stages of ARES.
However, some virtual outliers are synthesized within the ID region, which might have a high likelihood.

The effect of each stage is also confirmed in Table \ref{tab:ablationstage}. In the table, we select CIFAR-10 as ID and LSUN-C as OOD datasets; and we report FPR95.
FPR95 is the worst with 11.25\% when the Escape stage is excluded, as shown in the first row of Table \ref{tab:ablationstage}. If we exclude Expansion stage instead, the performance improves to 7.05\% as shown in the second row, but it is still not the best performer. Next, the third row shows further improved performance of 6.75\% by excluding Estimation stage instead.
Finally, our findings indicate that ARES shows the best performance of 6.00\% when all three main stages are harmoniously applied.
We conclude from the table that the performance of OOD detection is affected by each stage in the order of Escape, Expansion, and Estimation.

\begin{table}
  \caption{FPR95$\downarrow$ of loss ablation studies on CIFAR-10 as ID and LSUN-C as OOD. Values are percentages. The best performance is represented as bold.}
  \label{tab:ablationloss}
  \resizebox{\columnwidth}{!}{
  {\tiny
  \begin{tabular}{ccccc}
    \toprule
    \multirow{2}{*}{\textbf{Methods}} & \multicolumn{3}{c}{\textbf{Loss}} & \multirow{2}{*}{\textbf{LSUN-C}} \\
    \cmidrule(lr){2-4} & \textbf{NCE} & \textbf{CE} & \textbf{JSD} & \\
    \midrule
    ARES w/ NCE & \checkmark & - & - & 8.75 \\
    \midrule
    ARES w/ CE & - & \checkmark & - & 6.72 \\
    \midrule
    ARES w/ JSD (ours) & - & - & \checkmark & \textbf{6.12} \\
    \bottomrule
  \end{tabular}
  }
  }
  \vskip -0.4cm
\end{table}

\subsubsection{\textbf{Ablation on Discrimination Loss}} \label{abl:loss}
Another contributing factor of ARES is the discrimination loss of Eq. \eqref{eq:jsdloss} of the Divergence stage, i.e., loss for discriminating the energy-based score between inliers and virtual outliers. VOS used a non-linear logistic regression function and cross-entropy loss; whereas ARES employed Jensen-Shannon Divergence (JSD) loss to effectively model repulsion and maximize the distributional difference between ID and OOD datasets.

To observe the effectiveness of JSD loss, we compare it with two other losses in Table \ref{tab:ablationloss}; which are 1) the noise contrastive estimation (NCE) loss used in Contrastive Learning \cite{Sim_CLR} by calculating similarity, and 2) cross-entropy (CE) loss used in VOS. Both NCE loss and CE loss adopt logistic regression; while JSD loss directly discriminates energy-based scores without logistic regression.
In the table, we also selected CIFAR-10 as ID and LSUN-C as OOD datasets to report FPR95.
ARES with JSD loss, which is shown in the last row, shows the most outstanding performance of 6.12\%.
When adopting CE loss, the performance is 6.72\% as shown in the second row. With NCE loss, the performance further degrades to 8.75\% as shown in the first row.
This demonstrates that JSD loss is the most effective for modeling repulsion to maximize the distributional difference between ID and OOD using an energy-based score.
It should be noted that although the other two losses resulted in lower performances compared to the JSD loss, the differences were not significant and rather outperformed the baselines in Table \ref{tab:main_results}. This indicates the robustness of ARES in terms of sensitivity to the loss selection. Also, we can confirm the effectiveness of the initial three stages of ARES in that Table \ref{tab:ablationloss} is conducted with those stages involved.

\subsubsection{\textbf{Ablation on Training Epoch}} \label{abl:epoch}
Most of the baseline OOD detection methods presented in Table \ref{tab:main_results} were conducted with a total of 100 epochs of training, following the instructions in each paper. In contrast, ARES achieves superior results with a total training of 500 epochs, where a pre-training phase of 200 epochs for virtual outlier estimation is involved.
For a fair comparison, we decreased the total training epoch of ARES into the same level of baselines and reported FPR95 in Table \ref{tab:ablationepoch}, by selecting CIFAR-10 as ID dataset.
As shown in the table, ARES also demonstrated outstanding performance with both 100 and 200 total training epochs; the averaged FPR95 was 13.51\% with 100 epochs and 9.42\% with 200 epochs. Both results still outperform the second-best baseline, which is NPOS with 19.20\% as shown in Table \ref{tab:cifar10}.
This indicates that ARES can ensure its performance even with standard experimental settings which limit the training epochs, representing the robustness of ARES with regard to sensitivity to training cost.

\begin{table}[h]
  \caption{FPR95$\downarrow$ of training epoch ablation studies on CIFAR-10 as ID, by varying pre-training epochs and total training epochs. Values are percentages. The best performance is represented as bold.}
  \label{tab:ablationepoch}
  \resizebox{\columnwidth}{!}
  {
  \begin{tabular}{@{}lccccccc@{}}
    \toprule
    \multirow{2}{*}{\textbf{Pre/Total}}
    & \multicolumn{6}{c}{\textbf{OOD Datasets}} \\
    \cmidrule{2-7}
    \multicolumn{1}{c}{\textbf{Epochs}}
    & \textbf{Texture}
    & \textbf{SVHN}
    & \textbf{LSUN-C}
    & \textbf{iSUN}
    & \textbf{Place365}
    & \textbf{Average} \\
    \midrule
    \; 40/100 & 9.70 & 8.70 & 12.65 & 5.20 & 31.30 & 13.51 \\
    \; 80/200 & 7.25 & 3.95 & 5.90 & 4.25 & 25.75 & 9.42 \\
    \; 200/500 & \textbf{5.35} & \textbf{5.85} & \textbf{4.80} & \textbf{4.00} & \textbf{22.35} & \textbf{8.47} \\
    \bottomrule
  \end{tabular}
  }
  \vskip -0.2cm
\end{table}

\subsubsection{\textbf{Comparison of Time Complexity}} \label{abl:complexity}
\begin{figure}[h]
  \centering
  \begin{subfigure}{0.23\textwidth} 
    \includegraphics[width=\textwidth]{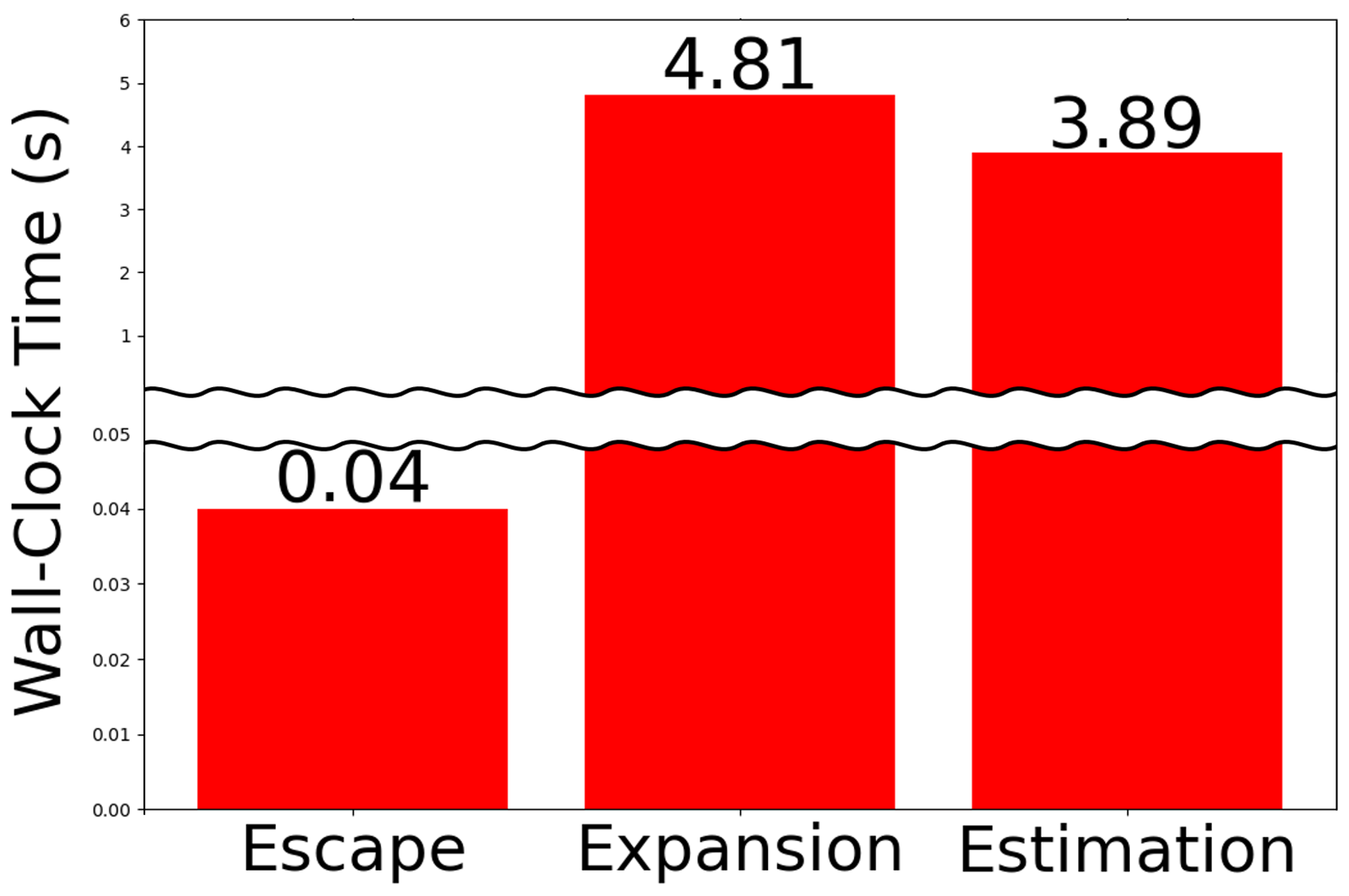}
    \captionsetup{justification=centering}
    \caption{Wall-clock time complexity \\ of stage ablation studies}
    \label{fig:abl1_time}
  \end{subfigure}
  \hfill
  \begin{subfigure}{0.23\textwidth}
    \includegraphics[width=\textwidth]{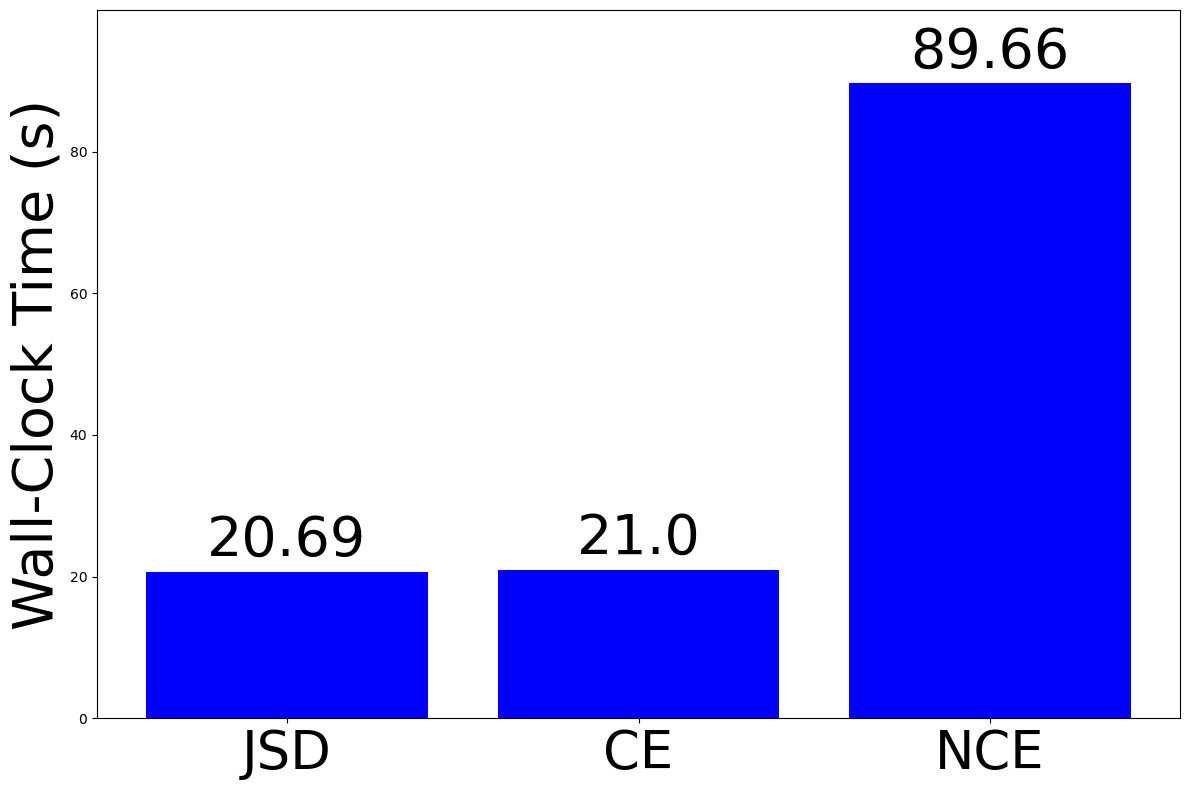}
    \captionsetup{justification=centering}
    \caption{Wall-clock time complexity \\ of loss ablation studies}
    \label{fig:abl2_time}
  \end{subfigure}
  \caption{Wall-clock time complexity of stage ablation and loss ablation studies on CIFAR-10. Wall-clock time was measured for one epoch in seconds.}
  \label{fig:time_complexity}
\end{figure}

At last, we analyze the time complexity of each stage of ARES in Figure \ref{fig:time_complexity}. The time complexity in the figure is measured with wall-clock time in seconds, and we report the duration for one epoch.
Figure \ref{fig:abl1_time} compares the complexity of Escape, Expansion, and Estimation stage. We can observe that the Escape stage takes the least complexity but other stages still fall within acceptable ranges.
In Figure \ref{fig:abl2_time}, the time complexity when adopting different losses can be observed. NCE loss takes the longest duration due to the matrix multiplication for the energy-based scores that are concatenated. JSD loss shows a similar duration to CE loss. We conjecture the estimation of Gaussian distribution in Eq. \eqref{eq:mu_jsd}--\eqref{eq:kld_m} was the bottleneck. However, such estimation in JSD loss resulted in an effective discrimination between inliers and virtual outliers, which leads to outperforming CE loss that was confirmed in Table \ref{tab:ablationloss}.

\section{Conclusion}  \label{sec:Conclusion}
In this paper, we propose a novel out-of-distribution detection framework called ARES. By addressing the limitation of conventional generation-based methods which generate outliers only within the ID region, ARES generates virtual outliers over a wide range beyond the ID region and its boundary through Escape, Expansion, and Estimation stages. At last, by utilizing JSD loss at Divergence stage, ARES achieves effective repulsion of energy-based scores, ultimately demonstrating improved OOD detection performance.
Quantitative and qualitative analyses confirm that ARES successfully synthesizes more valuable OOD-like virtual instances.
It should be noted that each stage can be applied to other baselines if needed, which would improve the baselines. This suggests the generalizability of ARES and we leave it as our future works.
We hope our research inspires future studies related to OOD detection in real industrial environments.



\bibliographystyle{ACM-Reference-Format}
\bibliography{sample-base}









\end{document}